\documentclass[letterpaper]{article} 
\usepackage{aaai25}  
\usepackage{times}  
\usepackage{helvet}  
\usepackage{courier}  
\usepackage[hyphens]{url}  
\usepackage{graphicx} 
\usepackage{natbib}  
\usepackage{caption} 
\usepackage{algorithm}
\usepackage{algorithmic}
\usepackage{xcolor} 
\usepackage{bbding}
\usepackage{pifont}
\usepackage{wasysym}
\usepackage{amssymb}
\usepackage{colortbl}
\usepackage{csquotes}
\usepackage{multirow}
\usepackage{amssymb}
\usepackage{dsfont}
\usepackage{diagbox}

\usepackage{dutchcal}
\usepackage{newfloat}
\usepackage{listings}
\DeclareCaptionStyle{ruled}{labelfont=normalfont,labelsep=colon,strut=off} 
\floatstyle{ruled}
\newfloat{listing}{tb}{lst}{}
\floatname{listing}{Listing}
\title{H-MBA: Hierarchical MamBa Adaptation for Multi-Modal Video Understanding in Autonomous Driving}
\author{
    Siran Chen\textsuperscript{\rm 1,\rm 2},
    Yuxiao Luo \textsuperscript{\rm 1,\rm 5},
    Yue Ma \textsuperscript{\rm 4},
    Yu Qiao\textsuperscript{\rm 1, \rm 3},
    Yali Wang\textsuperscript{\rm 1, \rm3,} \thanks{Corresponding author}
}
\affiliations{
    \textsuperscript{\rm 1} Shenzhen Institutes of Advanced Technology, Chinese Academy of Sciences, Shenzhen, China\\
    \textsuperscript{\rm 2} School of Artificial Intelligence, University of Chinese Academy of Science, Beijing, China \\
    \textsuperscript{\rm 3} Shanghai Artificial Intelligence Laboratory, Shanghai, China \\
    \textsuperscript{\rm 4} The Hong Kong University of Science and Technology, Hong Kong, China\\
    \textsuperscript{\rm 5} The Hong Kong Polytechnic University, Hong Kong, China \\
    chensiran17@mails.ucas.ac.cn, luoyuxiao@polyu.connect.hk,
    mayuefighting@gmail.com, \\
    \{yu.qiao, yl.wang\}@siat.ac.cn
}

\usepackage{bibentry}

\begin{document}

\maketitle

\begin{abstract}
With the prevalence of Multimodal Large Language Models(MLLMs), autonomous driving has encountered new opportunities and challenges. 
In particular, multi-modal video understanding is critical to interactively analyze what will happen in the procedure of autonomous driving.
However, videos in such a dynamical scene that often contains complex spatial-temporal movements,
which restricts the generalization capacity of the existing MLLMs in this field.
To bridge the gap, 
we propose a novel \textbf{H}ierarchical \textbf{M}am\textbf{ba} Adaptation (H-MBA) framework to fit the complicated motion changes in autonomous driving videos.
Specifically, 
our H-MBA consists of two distinct modules,
including Context Mamba (C-Mamba) and Query Mamba (Q-Mamba).
First,
C-Mamba contains various types of structure state space models,
which can effectively capture multi-granularity video context for different temporal resolutions.
Second,
Q-Mamba flexibly transforms the current frame as the learnable query, 
and attentively selects multi-granularity video context into query.
Consequently,
it can adaptively integrate all the video contexts of multi-scale temporal resolutions to enhance video understanding.
Via a plug-and-play paradigm in MLLMs,
our H-MBA shows the remarkable performance on multi-modal video tasks in autonomous driving,
e.g., for risk object detection, 
it outperforms the previous SOTA method with 5.5\% mIoU improvement.
\end{abstract}



\section{Introduction}

\begin{figure*}[t] 
    \centering
    \setlength{\abovecaptionskip}{0.1cm}
    \includegraphics[width=\textwidth]{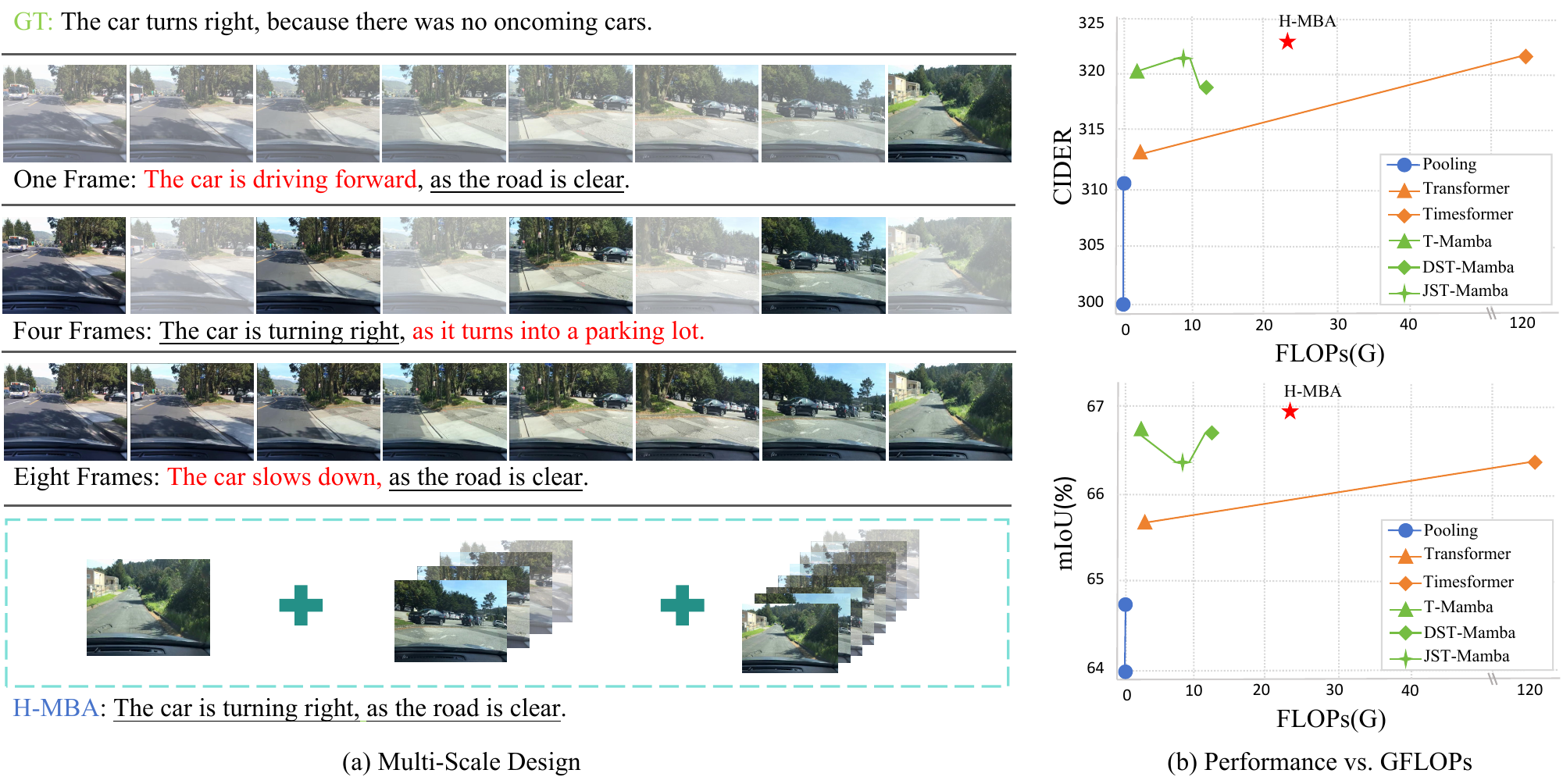}
    \caption{\textbf{Motivation.} (a) Previous models fail to give the correct description and justification with only single scale video input, while H-MBA combines multi-scale features and gets the most appropriate answer. (b) Performance of risk localization on Drama (Up/Down: Caption/Detection). Mamba blocks show advantage in the trade-off of performance and computation compared with attention modules, and our H-MBA achieves the best balance.  }
    \label{fig:motivation}
\end{figure*}

With the rapid advancement of modern artificial intelligence technologies, autonomous driving (AD) has made significant progress. 
Traditional end-to-end autonomous driving relies on precise environmental perception to make safe predictions and planning, 
Although it can directly generate planning routes or control signals from raw sensor data,
this process resembles a black box that excludes human drivers, making them difficult to understand the driving process and interact with the driving system.
To address the challenges, numerous studies have introduced datasets and methods~\cite{kim2018textual,deruyttere2019talk2car,kim2019grounding,atakishiyev2021explainable,malla2023drama,jin2023adapt} to incorporate natural language interaction and interpretability. 
Equipped with language models (i.e. BERT~\cite{devlin2018bert}), driving sensor data could be translated into natural language prompts to reveal how the models understand driving scenes.
However, these methods can only respond to predefined questions due to the limitations of language models, failing to work when faced with open-domain questions.
Multi-modal large language models (MLLMs) with extensive general knowledge and reasoning capabilities have been regarded as the future direction for such problems~\cite{zhou2023vision}, 
offering the potential as AI agent for AD tasks.

However,
the existing MLLM approaches~\cite{ding2023hilm,xu2023drivegpt4,fu2024drive} are limited to model complex temporal characteristics in the first perspective driving videos,
e.g., 
background changes,
motions of vehicles and pedestrians,  
and traffic signals.
Hence,
they usually fail in multi-modal video understanding in AD tasks,
for example, video MLLMs often get low performance in risk object localization~\cite{ding2023hilm}.
To solve the above limitation, 
we introduce a novel Hierarchical MamBa Adaptation (H-MBA) paradigm,
which can effectively and efficiently adapt MLLMs with hierarchical mamba modeling for video understanding in complex driving scenarios.

Specifically,
H-MBA consists of two distinct modules, 
Context Mamba (C-Mamba) 
and 
Query Mamba (Q-Mamba).
C-Mamba can flexibly learn video contexts by developing multi-granularity mamba models in multi-scale temporal branches.
In particular,
we introduce the low and high temporal resolution branches to model appearance and motion contexts in the driving videos, 
as shown in Fig~\ref{fig:motivation}(a), 
the low temporal resolution branch captures more obvious motion changes,
while the high resolution branch provides additional details. 
To master the diversified details in these ego-like videos,
With both of them, we get more comprehensive understanding of the video.
we further leverage three mamba structures for each temporal branch,
namely T-Mamba, DST-Mamba and JST-Mamba, 
to respectively learn Time-only, Divided Space-Time and Joint Space-Time granularity contexts.
After obtaining all the contexts,
Q-Mamba generates a learnable query from the current frame,
and attentively integrates multi-granularity contexts into this query,
via performing latent mamba in both temporal branches.
In this case,
the query could adaptively leverage rich video contexts within MLLMs to enhance the temporal understanding of driving videos.

To sum up,
there are three contributions in our paper.
First,
we are the pioneer to develop a mamba paradigm for hierarchical video understanding in the driving scenarios.
Compared to attention of video modeling~\cite{bertasius2021space},
our distinct mamba design exhibits the preferable performance with high computation efficiency in AD tasks,
e.g.,
our H-MBA achieves higher performance in the caption and detection tasks, but only costs about one-fifth of computation FLOPs compared to Timesformer, 
as shown in Fig~\ref{fig:motivation}(b).
Second,
our H-MBA is a plug-and-play video adaption module for MLLMs.
Taking advantage of novel C-Mamba and Q-Mamba designs,
it can flexibly guide MLLMs to learn driving video representations,
by adaptive extraction and integration of multi-scale multi-granularity contexts.
Finally, 
we conduct our H-MBA on multi-modal video understanding benchmarks in autonomous driving,
including DRAMA~\cite{malla2023drama} and BDD-X~\cite{kim2018textual}.
The extensive results show that,
our H-MBA achieves the state-of-the-art performance,
e.g.,
it gets 66.9\% mIoU on risk localization, with 5.5\% improvement compared with the previous SOTA approach~\cite{malla2023drama}.

\section{Related Work}
\paragraph{Multimodal Large Language Models} 
With the significant success of Large Language Models(LLMs)~\cite{brown2020language,chowdhery2023palm,devlin2018bert,wei2021finetuned,lin2024mope},
there are increasing research interest to expand the ability of LLMs to deal with multi-modal inputs and tasks.
Flamingo~\cite{alayrac2022flamingo} and PaLM-E~\cite{driess2023palm} seamlessly fuse image and text on massive image-text pairs, achieving great performance breakthrough among various multi-modal tasks;
BLIP-2~\cite{li2023blip} uses a lightweight Q-Former to bridge the gap of image and text modalities with little parameters and paves the way for LLaVA~\cite{liu2024visual}, 
MiniGPT-4~\cite{zhu2023minigpt}, InstructBLIP~\cite{dai2024instructblip};
Then the exploration extends into video domain, VideoChat~\cite{li2023videochat}, VideoChatGPT~\cite{maaz2023video} and Video-LLaMA~\cite{zhang2023video} further train their temporal module with video instruction-tuning data.
These models take multi-modal inputs and have conversations with users in multiple rounds, showing some basic logical reasoning ability. 
Moreover, several works, e.g. Shikra~\cite{chen2023shikra}, ContextDET~\cite{zang2023contextual} endow MLLMs with perceptual capabilities, enabling them to output bounding boxes, but the detection is just on image level.
While our H-MBA enhances Shikra with the proposed H-Mamba, could not only understand driving video inputs with basic captions,  but also provide bounding boxes of the risky object.

\begin{figure*}[t]
\centering
\setlength{\abovecaptionskip}{0.1cm}
\includegraphics[width=0.9\textwidth]{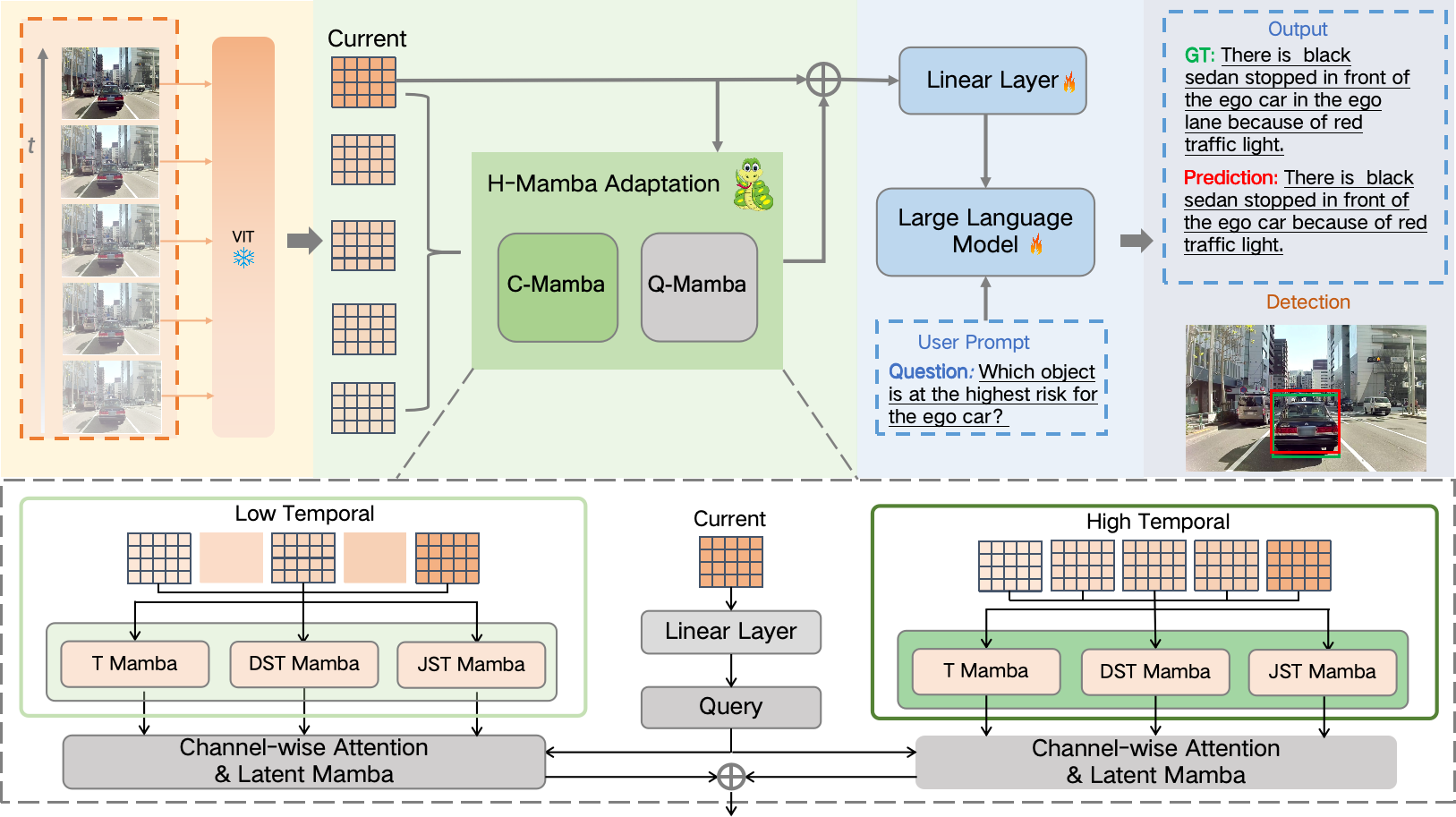}    
    \caption{\textbf{Pipeline of H-MBA framework.} We design an extra H-Mamba Adaptation block to process video input, the hierarchical refers to high and low temporal resolution and different Mamba-style modules here. After the fuse of Q-Mamba adapter, the multi-scale features are aligned with text query prompt and sent to the LLM to get the final answer.  }
\label{fig:overview}
\end{figure*}

\paragraph{State Space Model Series} 
The State Space Models (SSMs) have attracted widespread attention due to their high efficiency on modeling the dynamics and dependencies of long-term language sequences,
S4~\cite{gu2021efficiently} designs a structured state-space sequence model especially for long-range dependencies, with only liner complexity, 
and followed by various variants~\cite{smith2022simplified,dao2023hungry}.
Furthermore, Mamba~\cite{gu2023mamba} with data-dependent SSM layer and parallel scan selection mechanism (S6) surpasses transformers in long sequence tasks.
For the vision domain, Vision Mamba~\cite{zhu2024vision} and VMamba~\cite{vmamba} utilize different scan mechanisms for 2D image processing,
ViVim~\cite{yang2024vivim} and VideoMamba~\cite{li2024videomamba} expand the scan for 3D videos, these Mamba blocks show advantage in effectiveness and efficiency compared to transformer-based modules.

\paragraph{MLLMs for Autonomous Driving.} 

Traditional end-to-end autonomous driving exclude human drivers from involvement in the driving process like black boxes, making it hard for drivers to understand and interact with the system, though some works~\cite{wang2021learning,kim2018textual,deruyttere2019talk2car,jin2023adapt,malla2023drama} aim to interpret vehicle status from sensor inputs in natural language, they can only solve predefined questions with poor generality.
MLLMs provide new directions to address such problems, in fact, they have been widely applied in various downstream tasks as AI agent~\cite{luo2023valley,liang2023code,li2024llava,karabacak2023embracing}.
DriveLikeHuman~\cite{fu2024drive} conducts simulation to prove the potential of LLM to make human-like decisions. DriveGPT4~\cite{xu2023drivegpt4} is fine-tuned on large GPT-generated conversations to enable models to respond to user queries.
HiLM-D~\cite{ding2023hilm} leverages high and low resolution branches to let models put more attention on the location of small risky objects.
However, they merely utilize existing temporal processing methods while ignoring the characteristics of driving videos,
for instance, HiLM-D uses ST-Adapter~\cite{pan2022st} to process video inputs, though getting higher caption performance, the detection results show a decline than the image inputs.
Our method not only allows user queries as prompts to perform various tasks, 
but also incorporates multi-scale/granularity paradigm, which enables more comprehensive understanding for driving videos and boosts performance on the tasks.

\begin{figure*}[t]
\centering
\setlength{\abovecaptionskip}{0.1cm}
\includegraphics[width=0.8\linewidth]{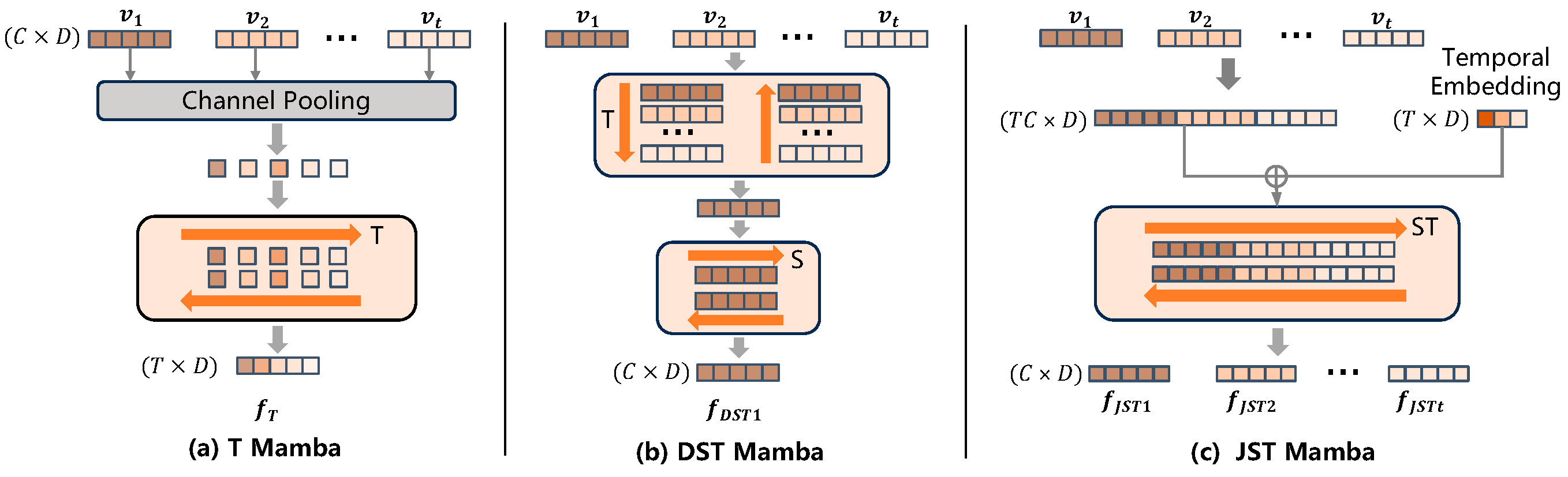}    
    \caption{\textbf{Illustration of three different space-time Mamba sequences in our C-Mamba.} So we could get multi-granularity video features to fit for various tasks.
     }
\label{fig:mamba}
\end{figure*}

\section{Method}

\subsection{Overview}
The framework of our model is presented in  Fig~\ref{fig:overview}. 
We choose Shikra~\cite{chen2023shikra} pre-trained with large location-aware data as the baseline model, 
which is able to give the coordinates of the output object, in addition to the basic QA ability.
The structure is straightforward and simple, 
consisting of pre-trained CLIP~\cite{radford2021learning} ViT-L/14 as the frozen visual encoder, 
Vicuna-7/13B as the LLM and a fully connected layer for feature transition and alignment.
The original Shikra can only handle image input, it's not enough for the driving scenarios where the understanding of temporal information is very important.
While preserving the baseline structure, we expand the input phase to take multi-frame video inputs, 
then leverage the H-MBA to model the spatio-temporal representation.
H-MBA consists of Context Mamba (C-Mamba) to capture multi-granularity video context and Query Mamba (Q-Mamba) to adaptively learn and fuse these features, they will be introduced in detail later.
The processed features are added to the original current frame in a residual manner to preserve the pre-training performance. 
Users could flexibly give query prompts for different tasks, e.g., \enquote{What's the ego car's action in the video and explain the reasons.}, \enquote{Which object is at the highest risk?}.
Then the pre-trained LLM receives the multi-modal inputs and gives the corresponding answer.

\subsection{C-Mamba: Context Mamba for Multi-Granularity Video Modeling} 
To better capture the object motion and model background changes in complex driving scenarios, we use a hierarchical C-Mamba to process multi-frame video context. 
The \enquote{hierarchical} has two meanings here.
Firstly, 
considering that speed serves as the \enquote{bridge} connecting spatio-temporal changes, 
the driving video context is highly correlated to the speed of the ego car, 
for example, the ego car may stop, drive quickly or turn to another lane, 
the video contexts vary a lot correspondingly.
Thus we introduce the paradigm that incorporates both high and low temporal resolutions to fit the diverse speed dynamics,
the low branch observes more obvious motion changes,
while the high branch involves more detailed information.
Secondly, for each temporal resolution branch,
we utilize different Mamba state space architectures to learn multi-granularity spatio-temporal video features.

To be more specific,
we design three Mamba-based modules,
i.e., temporal only sequence Mamba, divided space-time sequence Mamba and joint space-time sequence Mamba, each block learns specific ST patterns.
Note that we leverage bidirectional Mamba (Bi-Mamba) to model vision-specific tasks here, which adapts another flattened visual sequences through simultaneous forward and backward SSMs.
We label the frame features from frozen visual encoder as $\mathbf{v}_1, \mathbf{v}_2, ..., \mathbf{v}_t \in \mathbb{R}^{C \times D}$, $t$ is the frame number, $C$ is the number of patch channel per frame, and $D$ is the feature dimension. 
The three distinct structures could be seen in Fig~\ref{fig:mamba}, and introduced below.

\paragraph{T Mamba.}
For temporal only sequence Mamba, it first pools the patch dimension of each frame feature to get the overall frame representation, then the $t$ frame features are sent to the bidirectional mamba block to learn the temporal relation, this module is suitable for modeling the overall changes of the background:
\begin{equation}
\mathbf{f_T} = \mathbf{Mamba}(pooling(\mathbf{v}_1,\mathbf{v}_2,...,\mathbf{v}_t)) 
\label{eq:t}
\end{equation}

\paragraph{DST Mamba.}
For the divided space-time sequence Mamba (DST), 
we first separately process the temporal sequence for each patch channel $\mathbf{v}_i^c$ with Mamba block, so the patches could involve motion change information of the local part, labeled as $\mathbf{v_T}$. 
Then the processed patch sequences of the $i_{th}$ frame $\mathbf{v_T}_i$ are sent to spatial Mamba block and get the divided space-time features:
\begin{equation}
\mathbf{f_{DST}} = \mathbf{Mamba}(\mathbf{v_T})
\end{equation}
\begin{equation}
\mathbf{v_T}^c = \mathbf{Mamba}(\mathbf{v}_1^c,\mathbf{v}_2^c,...,\mathbf{v}_t^c)  
\label{eq:divided}
\end{equation}

\paragraph{JST Mamba.}
For the joint space-time sequence Mamba (JST), 
thanks to the $O(n)$ computational complexity of Mamba algorithm, 
we are able to jointly process all the patch sequences, which is computationally extravagant for transformer-based method. 
The t frame patch feature sequences are first concatenated, we add a learnable temporal embedding($\mathbf{TE}$) for each frame to distinguish temporal differences.
The $t \times C$ sequences could be sent to Mamba to jointly learn the global spatio-temporal relation.
\begin{equation}
 \mathbf{f_{JST}} = \mathbf{Mamba}(Concat(\mathbf{v}_1,\mathbf{v}_2,...,\mathbf{v}_t) + \mathbf{TE}) 
\label{eq:joint}
\end{equation}

\begin{table*}[t]

\centering
\resizebox{0.8\linewidth}{!}{ 
  \begin{tabular}{l|c|ccccc|c} 
    \hline
     \textbf{Method} & \textbf{MLLM} & \textbf{B4} $\uparrow$ & \textbf{M} $\uparrow$ & \textbf{R} $\uparrow$ & \textbf{C} $\uparrow$ & \textbf{S} $\uparrow$ & \textbf{mIoU} $\uparrow$\\
    \hline
     \textcolor{gray}{SAT~\cite{xu2015show}} & \textcolor{gray}{\XSolidBrush}  & \textcolor{gray}{0.531} &\textcolor{gray}{ 0.386} & \textcolor{gray}{0.694} & \textcolor{gray}{3.583} & \textcolor{gray}{0.544} & -\\
     
     \textcolor{gray}{ResNet-101~\cite{he2016deep}} & \textcolor{gray}{\XSolidBrush} & - & -& -& - & - & \textcolor{gray}{60.0} \\
     
     \textcolor{gray}{LCP~\cite{malla2023drama}} & \textcolor{gray}{\XSolidBrush}  & \textcolor{gray}{0.520} & \textcolor{gray}{0.379} & \textcolor{gray}{0.688} & \textcolor{gray}{3.501} & \textcolor{gray}{0.526} & \textcolor{gray}{59.7} \\
     
     \textcolor{gray}{LCP(with OF)~\cite{malla2023drama}} & \textcolor{gray}{\XSolidBrush} & \textcolor{gray}{0.547} & \textcolor{gray}{0.391} & \textcolor{gray}{0.700} & \textcolor{gray}{3.724} & \textcolor{gray}{0.560} & \textcolor{gray}{61.4}\\
    \hline    
    BLIP-2~\cite{li2023blip}  & \Checkmark & 0.506 & 0.386 & 0.651 & 2.585 & 0.537 & 46.3 \\
     LLaVA~\cite{li2024llava} & \Checkmark & -& -& -& - & -& 47.2 \\
     Shikra~\cite{chen2023shikra} & \Checkmark  & 0.513 & 0.379 & 0.677 & 3.002 & 0.539 & 64.0\\
     Video-Chat~\cite{li2023videochat} & \Checkmark & 0.368 & 0.290 & 0.482 & 1.530 & 0.435 & - \\  

     HiLM-D(Video)~\cite{ding2023hilm} & \Checkmark & - & -& -& - & - & 59.2 \\
     
     Video-LLaMA~\cite{zhang2023video} & \Checkmark & 0.481 & 0.384 & 0.640 & 2.135 & 0.507 & 42.8 \\
     
     H-MBA & \Checkmark & \textbf{0.536} & \textbf{0.394} & \textbf{0.697} & \textbf{3.227} & \textbf{0.558} & \textbf{66.9} \\
    \hline  
 \end{tabular}
}
\caption{\textbf{Performance comparison on DRAMA.}  Note that we use the detection results from HiLM-D~\cite{ding2023hilm}, but its caption results are conducted on their extra private labeled data, we report the caption results of the original data.
Our H-MBA gets the highest risk detection result, achieving 66.9\% mIoU, with 5.5\% improvement than previous SOTA, LCP(with OF), OF denotes optical flow.
}
\label{tab:drama}      
\end{table*}

\subsection{Q-Mamba: Query Mamba for Multi-Scale Video Context Adaptation}
After obtaining the multi-granularity features, another important question is how to fuse these extra knowledge to the pre-trained MLLM. 
We have three kinds of designs here.
The first straightforward and simple way is to directly add all the weighted learned features to the original visual features, 
we name it directly add (DA) adapter as the baseline method.
Then, motivated by the inception~\cite{szegedy2015going} network which processes features with different convolution kernels and concatenate them in parallel, we also use such paradigm and concat all the features in the feature dimension D and get $\mathbf{f}_{cat} \in \mathbb{R}^{C \times (n*D)}$, 
n is the number of different granularity features.
After that, we use a linear-layer to transform the features form $n*D$ to $D$ , and this format is called inception concat (IC) adapter.
\begin{equation}
\mathbf{F_{IC}} = \mathbf{FC}(Concat(\mathbf{f}_1,...,\mathbf{f}_n)) 
\label{eq:IC}
\end{equation}

Furthermore, we design a Query Mamba (Q-Mamba) adapter. 
Considering that the visual encoder is frozen during fine-tuning, 
the original frame visual feature from the encoder is aligned with LLM and plays a vital role in the pre-training and inference, 
in order to preserve the performance of the original model as much as possible, 
we flexibly transform the current frame feature as query $\mathbf{Q}^c$ for each patch channel, 
and the multi-scale video context features are treated as keys and values, 
so each channel-wise query $\mathbf{Q}^c$ could adaptively learn what's valuable from the multi-level features of the patch.
This ensures that the model could learn the most appropriate features for different tasks through a unified framework.
Then a latent mamba is used to perform channel interaction and learn an overall understanding.
We call this Q-Mamba,
the process is perceiver~\cite{jaegle2021perceiver} style, 
and can be formulated as:
\begin{equation}
\mathbf{F_Q}^c = \sum\limits_{i=1}^{n}\mathbf{Mamba}(\mathbf{CrossAttn}(\mathbf{Q}^c, \mathbf{f}_i^c, \mathbf{f}_i^c)) 
\label{eq:CW}
\end{equation}

\section{Experiments}
\label{headings}

\begin{table*}[t]
\small

\centering
\resizebox{0.95\linewidth}{!}{

    \begin{tabular}{l|c|c|ccccc|ccccc} 
    \hline
     \multirow{2}{*}{\textbf{Method}}  & \multirow{2}{*}{\textbf{MLLM}} & \multirow{2}{*}{Signal}  &  \multicolumn{5}{c|}{Description} &  \multicolumn{5}{c}{Justification}           \\
    &\ &  & \textbf{B4} $\uparrow$ & \textbf{M} $\uparrow$ & \textbf{R} $\uparrow$ & \textbf{C} $\uparrow$ & \textbf{S} $\uparrow$  &\textbf{B4} $\uparrow$ & \textbf{M} $\uparrow$ & \textbf{R} $\uparrow$ & \textbf{C} $\uparrow$ & \textbf{S} $\uparrow$ \\
    \hline
  \textcolor{gray}{S2VT ~\cite{venugopalan2015sequence}} & \textcolor{gray}{\XSolidBrush} & \textcolor{gray}{\XSolidBrush} & \textcolor{gray}{0.302} & \textcolor{gray}{0.275} & -- & \textcolor{gray}{1.798} & -- & \textcolor{gray}{0.063} & \textcolor{gray}{0.112} & -- & \textcolor{gray}{0.534} & -- \\
     \textcolor{gray}{ADAPT-32~\cite{jin2023adapt}} & \textcolor{gray}{\XSolidBrush}   & \textcolor{gray}{\Checkmark}  & \textcolor{gray}{0.346} & \textcolor{gray}{0.306} & \textcolor{gray}{0.628} & \textcolor{gray}{2.475} & \textcolor{gray}{0.597} & \textcolor{gray}{0.114} & \textcolor{gray}{0.152} & \textcolor{gray}{0.320} & \textcolor{gray}{1.026} & 
    \textcolor{gray}{0.256} \\
  \textcolor{gray}{ADAPT-8~\cite{jin2023adapt}} & \textcolor{gray}{\XSolidBrush} & \textcolor{gray}{\Checkmark}  & \textcolor{gray}{0.329} & \textcolor{gray}{0.290} & \textcolor{gray}{0.609} & \textcolor{gray}{2.257} & \textcolor{gray}{0.579} & \textcolor{gray}{0.084} & \textcolor{gray}{0.134} & \textcolor{gray}{0.306} & \textcolor{gray}{0.837} & \textcolor{gray}{0.235} \\
   BLIP-2~\cite{li2023blip}  & \Checkmark & \XSolidBrush & 0.165 & 0.231 & 0.497 & 1.076 & 0.481 & 0.043 & 0.088 & 0.218 & 0.455 & 0.087 \\
    Video-Chat~\cite{li2023videochat} & \Checkmark & \XSolidBrush   & 0.145 & 0.211 & 0.473 & 0.947 & 0.451 & 0.038 & 0.080 & 0.166 & 0.402 & 0.073 \\
    Video-LLaMA~\cite{zhang2023video} & \Checkmark & \XSolidBrush & 0.168 & 0.238 & 0.509 & 1.045 & 0.486 & 0.045 & 0.086 & 0.221 & 0.464 & 0.101
    \\
   Shikra~\cite{chen2023shikra} & \Checkmark & \XSolidBrush  & 0.244 & 0.256 & 0.549 & 1.349& 0.521 & 0.072 & 0.125 & 0.288 & 0.709 & 0.209 \\
   H-MBA & \Checkmark & \XSolidBrush  & \textbf{0.301} & \textbf{0.285} & \textbf{0.601} & \textbf{1.996} & \textbf{0.562} & \textbf{0.088} & \textbf{0.137} & \textbf{0.308} & \textbf{0.949} & \textbf{0.238} \\
    \hline    
     \textcolor{gray}{ADAPT-32~\cite{jin2023adapt}} & \textcolor{gray}{\XSolidBrush} & \textcolor{gray}{\Checkmark} &  \textcolor{gray}{0.355} &  \textcolor{gray}{0.310} & -- &  \textcolor{gray}{2.589} & -- &  \textcolor{gray}{0.123} &  \textcolor{gray}{0.163} & -- &  \textcolor{gray}{1.161} & -- \\
    DriveGPT-4~\cite{xu2023drivegpt4} & \Checkmark & \Checkmark & 0.300 & 0.298 & -- & 2.140 & -- & 0.094 & 0.146 & -- & 1.027 & -- \\ 
    H-MBA & \Checkmark & \XSolidBrush & 0.310 & 0.296 & -- & 2.188 & -- & 0.098 & 0.148 & -- & 1.088 & -- \\
    \hline  
    \end{tabular}
    }
\caption{\textbf{Performance comparison on BDD-X}. The top part of the table is on full test set, and the bottom part is on 500 randomly sampled cases. We get the SOTA results among MLLM based models, we also have advantage for justification compared to ADAPT-8 even it takes extra signal.
}
\label{tab:bddx}      
\end{table*}

\paragraph{Datasets and Evaluation metrics.} DRAMA~\cite{malla2023drama} is a benchmark evaluating the visual reasoning of driving risks, meanwhile, it provides important object bounding boxes with captions describing their risk from the ego-car perspective. The whole dataset contains 17,785 two-second interactive driving scenarios, while considering spatio-temporal relationships from videos.
BDD-X~\cite{kim2018textual} is a driving-domain caption dataset, consisting of nearly 7000 videos that are collected from BDD100K, and the videos are manually captioned with vehicle behaviors, such as accelerating, and also accompanied with text justification for the behavior. The videos are divided into around 29,000 clips, and the clips' length duration ranges from 1s to 40s, so the processing of temporal information is critical.
We perform two tasks for the driving videos:(1) language caption, including the DRAMA and BDD-X datasets, the captioning performance is evaluated with standard metrics, such as BLEU-4(B4) , METEORP(M), ROGUE(R), CIDER(C) and SPICE(S). (2) Detection for DRAMA, the performance is evaluated with mean intersection over union (mIoU) for the prediction of bounding box.
\paragraph{Implementations details.} 
Following previous methods~\cite{chen2023shikra,li2023blip, li2023videochat}, we use the pre-trained Shikra~\cite{chen2023shikra} as the baseline, and finetune the model with task specific data. 
We uniformly sample $L$ frames for a video, the frames are resized and cropped to 224 $\times$ 224 resolution.
For DRAMA dataset, $L=5$, and we use the Spot-Captioning training format, which allows a bounding box along with the output caption.
And for BDD-X, the videos are longer, we set $L=8$ and use the Instruct-tuning format.
For other MLLM based methods, unless specifically mentioned, we follow the official training code with the same experimental settings. 
All the experiments are done with 4 A6000 GPUs, we train the model for 5 epochs with $2e^{-5}$ learning rate in cosine annealing schedule~\cite{loshchilov2016sgdr}.

\begin{table*}[t]
\small

\centering
\resizebox{0.9\linewidth}{!}{

  \begin{tabular}{l|l|c|c|cccc|cccc} 
    \hline
    \multicolumn{4}{c|}{\textbf{Dataset}}  & \multicolumn{4}{c}{\textbf{DRAMA}} & \multicolumn{4}{c}{\textbf{BDD-X}}  \\
    \hline
    \multirow{2}{*}{\textbf{Module}} &\multirow{2}{*}{\textbf{Paradigm}} &\multirow{2}{*}{\textbf{GFlops}}  &\multirow{2}{*}{\textbf{Params}}&
    \multicolumn{2}{c}{Caption} & \multicolumn{2}{c|}{Detection} & \multicolumn{2}{c}{Description} &  \multicolumn{2}{c}{Justification}    \\
    & & & &\textbf{B4} $\uparrow$  & \textbf{C} $\uparrow$ & \textbf{mIoU}$\uparrow$ & \textbf{Acc}$\uparrow$ & \textbf{B4} $\uparrow$  & \textbf{C} $\uparrow$ &\textbf{B4} $\uparrow$  & \textbf{C} $\uparrow$ \\
    \hline
  Image & None & -- & -- & 0.513 &  3.002 & 64.0 & 70.8 &  0.244  & 1.349  & 0.072 & 0.709  \\
  Average Pool  & None & -- & -- & 0.526 & 3.107 & 64.7 & 73.3 & 0.274 & 1.776  & 0.078  & 0.803 \\
  \hline
  Transformer& Time-only & 2.7 & 13.7M & 0.526 & 3.126 & 65.6 & 73.7 & 0.278   & 1.809  & 0.080  & 0.850  \\
  T Mamba & Time-only & 2.2 & 7.7M & 0.533 & 3.207 & 66.8 & 75.4 & 0.284 & 1.907 & 0.085 & 0.857 \\
  \hline
  Timesformer & S-T divided &116.9 & 71.4M & 0.533 & 3.218 & 66.3 & 74.3 & 0.285 & 1.898 & 0.082 & 0.852 \\
  DST Mamba &  S-T divided  & 12.0 & 14.4M & 0.527  & 3.187 & 66.6 & 75.0 & 0.283  & 1.883 & 0.087  & 0.947 \\
  JST Mamba & S-T Joint & 8.7 & 7.7M & 0.536  & 3.221 & 66.3 & 74.8 & 0.298  & 1.957 & 0.086  & 0.928 \\

     \hline  
    \end{tabular}
    }
\caption{\textbf{Ablation for different temporal information processing module}. Acc refers to the proportion that IoU\textgreater0.5.
}
\label{tab:module}      
\end{table*}

\begin{table*}[t]

\centering
\resizebox{0.7\textwidth}{!}{ 
    \begin{tabular}{l|cccc|cccc} 
    \hline
   \ & \multicolumn{4}{c|}{\textbf{DRAMA}} & \multicolumn{4}{c}{\textbf{BDD-X}}  \\
    \hline
    \multirow{2}{*}{\textbf{Adapter}}  & \multicolumn{2}{c}{Caption} &\multicolumn{2}{c|}{Detection} & \multicolumn{2}{c}{Description} &  \multicolumn{2}{c}{Justification}          \\
     & \textbf{B4} $\uparrow$  & \textbf{C} $\uparrow$ & \textbf{mIoU}$\uparrow$ & \textbf{Acc}$\uparrow$   &\textbf{B4} $\uparrow$ & \textbf{C} $\uparrow$ &\textbf{B4} $\uparrow$ & \textbf{C} $\uparrow$ \\
    \hline

    Direct Addition (DA) & 0.520  & 3.157 & 66.6 & 75.1 & 0.301  & 1.908 & 0.081  & 0.892
    \\
    Inception Concat (IC) & 0.515  & 3.027 & 64.3 & 72.5 & 0.246  & 1.461 & 0.072 & 0.712
    \\
    Q-Mamba & 0.536  & 3.227 & 66.9 & 75.6 & 0.301  & 1.996 & 0.088  & 0.949
    \\
   
    \hline  
    \end{tabular}
    }
\caption{\textbf{Ablation for the choice of adapter.} Q-Mamba keeps the optimal performance across multiple tasks.}
\label{adapter}      
\end{table*}

\subsection{SOTA Comparison}
We conduct experiments on DRAMA and BDD-X datasets and compare the results with both MLLM and non-MLLM based methods. 
Note that, the non-MLLM models are designed for specific tasks, 
limited to singular functionalities, the results are gray in the tables. 
In contrast, the MLLM-based approaches offer greater flexibility, enabling tailored responses based on user questions, all MLLMs are fine-tuned in same setting with official scripts,
and as a unified framework, we only use visual inputs for all the tasks.

The risk localization task on DRAMA requires the output of the object with highest risk for the ego car along with the bounding box, different from traditional vehicle object detection, the model should not only have detection and localization ability,
but also have a general understanding of the traffic environment and vehicle motion,
which is helpful for risk warning and improving driving safety.
As shown in Table~\ref{tab:drama},
for the caption scores of risk objects, our H-MBA obtains the optimal results among MLLM models, 
moreover, we also achieve the highest detection mIoU score, i.e., 66.9\%, with 5.5\% improvement than previous SOTA LCP~\cite{malla2023drama} even with optical flow. 
Among MLLM based methods, HiLM-D~\cite{ding2023hilm}, the previous SOTA, which uses ST-Adapter to process temporal information, but gets lower mIoU in video format than its image format, while our H-MBA is both helpful for the caption and detection performance, it shows the effectiveness. 

The understanding of BDD-X dataset requires models to describe the ego car's action and explain the intention. 
Many previous methods utilize extra control signals, such as driving speed and angle,
to help make the prediction.
To keep consistent with the former task, we just use 8 frames inputs as a uniform setting.
Our H-MBA gets the SOTA results among MLLM based models in Table~\ref{tab:bddx}, including BLIP-2~\cite{li2023blip}, Video-Chat~\cite{li2023videochat}, Video-LLaMA~\cite{zhang2023video} and DriveGPT-4~\cite{xu2023drivegpt4} with extra signal.
Directly using general video foundation models for AD tasks may not be superior due to domain bias, which also proves the effectiveness of our proposed method.
And for the previous non-MLLM SOTA ADAPT~\cite{jin2023adapt}, 
though the extra signal helps to get higher description scores, our H-MBA still gains better performance for the justification of the action, which means that MLLMs learn good understanding of the driving scenario.

\begin{figure*}[t]
\centering
\setlength{\abovecaptionskip}{0.1cm}
\includegraphics[width=0.82\linewidth]{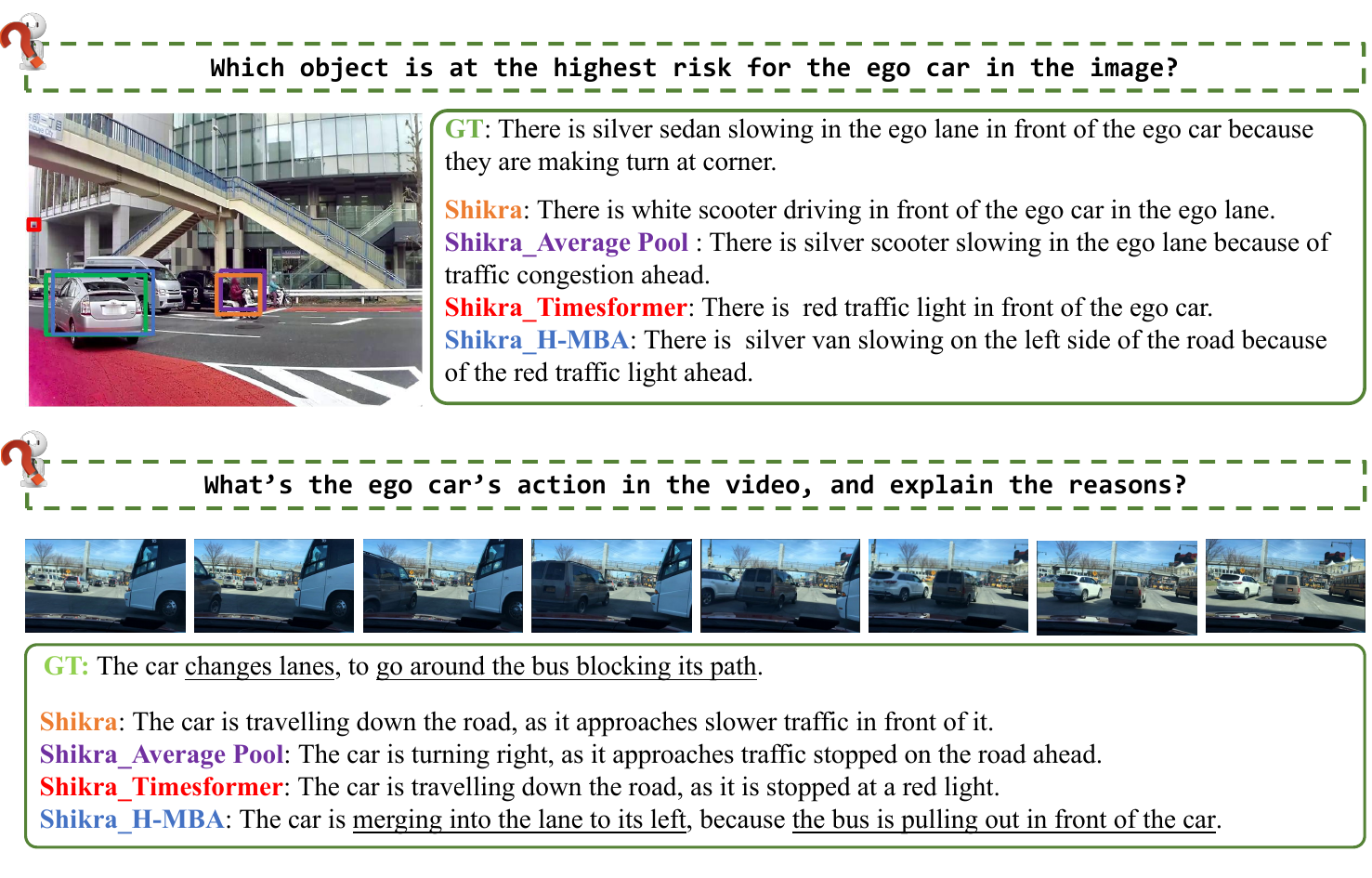}    
    \caption{\textbf{Visualization comparison for the output of different temporal processing modules.} For some rare scenarios, such as lane changing caused by a bus occupying the road ahead, our H-MBA can recognize the action and provide reasonable explanations, while others focus on the wrong points.  }

\label{fig:compare}
\end{figure*}

\subsection{Ablations}

\paragraph{Choice of Temporal processing module.} In this part, we first ablate which temporal module is suitable for driving tasks with both consideration of performance and computation cost. We compare our designed mamba-series modules with corresponding mainstream attention temporal modules in Table~\ref{tab:module}.
\enquote{Image} means we only use the current frame feature of the video, which is also the Shikra baseline.
And \enquote{Average Pool} means we mean pool the video features in the frame dimension,
the performance has slight improvement but still not satisfying without temporal sequence modeling.
Then we compare our T Mamba module with Transformer~\cite{vaswani2017attention} block, both of them directly learn the relationships between frames.
We can see that, Mamba block has advantages in both performance and Flops cost, in such small scale and lightweight modeling.
Next, we consider spatio-temporal divided modeling, corresponding to the Timesformer~\cite{bertasius2021space} and DST Mamba in the table. 
Though the spatial interaction further improves the results on these tasks, the extra computation cost of Timesformer is the highest. 
Compared to Timesformer, our Mamba-based methods seem to have more advantages in risk object detection, which needs to integrate temporal changes while retaining the current positions of the objects in current frame.
Finally, for our JST Mamba, we fully leverage the capability of Mamba block to process long sequences, 
with added temporal embedding to explicitly denote temporal distinctions. 
It has better performance in caption tasks. 
Note the corresponding attention-based operation requires excessive computational resources, which would cause Out of Memory (OOM) in our machine.
On the whole, the Mamba-based modules achieve higher performance with lower computation cost, while different Mamba granularities have advantages in different tasks, and our H-MBA adaptively assembles them to benefit all the tasks.

\paragraph{Choice of Adapter.}
Next we figure out the optimal setting of three feature adapters.
As shown in Table~\ref{adapter},
DA aggregates all the features by learnable weights, yet the results indicate that it merely yields a somewhat averaged outcome.
The results of IC are relatively low, it may be because the weights are difficult to converge for too many concated features.
Then for our Q-Mamba, 
the results have achieved optimal results on each metric, some even higher than single one,
which demonstrates the model could effectively learn the most relevant knowledge from the multi-scale features.
More ablations could be seen in the Appendix.

\paragraph{Visualization.}
We provide numerous visual examples and compare the results obtained using different time processing modules, as shown in Fig~\ref{fig:compare}. 
In addition to normal scenes like \enquote{the car is driving} and \enquote{the car is stopped}, our model can also accurately identify \enquote{turning} and \enquote{merging lane} behaviors. 
For some rare scenarios, such as lane changing caused by a bus occupying the road ahead in the example, 
the model can also recognize the action and provide reasonable explanations, 
while others focus on the wrong points, e.g., traffic lights. 
These examples vividly demonstrate the advantages of hierarchical feature processing.
Moreover, we show some real-world accident cases in the appendix, demonstrating the generalization of H-MBA for practical application to enhance driving safety.

\section{Conclusion}
Recent MLLM based approaches have primarily focused on data-labeling and fine-tuning for driving tasks, 
while we focus on the complicated scene changes in driving videos, and put forward H-MBA for multiple driving tasks.
H-MBA utilizes high and low temporal resolution branches, integrated with three different Mamba-based processors to learn features of different granularity.
The hierarchical features are fused by the Q-Mamba adapter and sent to LLMs to boost the performance when handling driving videos, with little extra computation.
We achieve SOTA results in several tasks such as risk object localization, demonstrating the effectiveness of our approach, and showing the potential for practical application and improving driving safety.

\section*{Acknowledgements}
This work was supported by the National Key R\&D Program of China(NO.2022ZD0160505), the National Natural Science Foundation of China under Grant(62272450), and the Joint Lab of CAS-HK.

\bibliography{aaai25.bib}

\end{document}


\appendix

\section{Additional Implementation Details}
\paragraph{Implementations.}
The original DRAMA\cite{malla2023drama} and BDD-X\cite{kim2018textual} datasets do not provide the suitable jsonl files for MLLM training, thus we follow the data format used in Shikra\cite{chen2023shikra}, and prepossess the raw data format by conducting video data cutting, format reorganization, and coordinate transformation, thereby generating files that meet the fine-tuning requirements. 
Additionally, we perform repairs and optimizations on unsuitable textual contents.
For the training process, it takes around 5 to 7 hours for 5 epochs, and for inference, the model processes approximately 2800 examples in around 10 minutes, with an average processing time of about 0.2 seconds per video. This demonstrates the practical potential of our method for real-world applications.
And for the parameter initialization, the three Mamba blocks in C-Mamba are randomly initialized with a normal distribution, while the parameters in Q-Mamba are initialized as identity matrix to gradually learn the knowledge and fuse the learned knowledge to the frame features.
Note that, for DRAMA we use the current frame as the frame features since we predict the risk object of the current moment,
and for BDD-X we use the pooling of all video frames as the frame features, because the understanding of the vehicle's behavior requires overall motion information.

\begin{figure}[b]
\centering
\includegraphics[width=\linewidth]{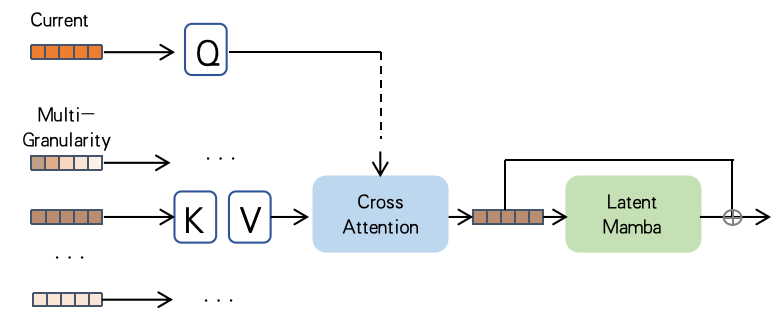}   
    \caption{\textbf{Mamba blocks for 1D sequence.}  }
\label{fig:mamba}
\end{figure}

\paragraph{Detailed Illustration of Q-Mamba}
Here we provide a more detailed illustration explanation of Q-mamba, which is not elaborated in the main text. 
After obtaining the multi-granularity features, we apply cross attention between each feature and the original image features along the channel dimension.
The current frame feature is transformed to query by a FC layer, and the multi-granularity features are regarded as keys and values directly.
The resulting features are then fed into the latent Mamba layer to enable interactions across the channel dimension. 
The latent Mamba layer is shared, and after computation, 
the output is combined with the original features through a residual connection.

\section{Preliminaries: State Space Models}
State Space Models (SSMs)\cite{gu2023mamba} are conceptualized based on continuous systems that map a 1D function or sequence, $x(t) \in \mathbb{R} \rightarrow y(t) \in \mathbb{R}$,
through a hidden state $h(t) \in \mathbb{R}^N$.
As linear time-invariant systems, the process is formally formulated as linear ordinary differential equation(ODE) to model the relations, 
where $\mathbf{A} \in \mathbb{R}^{N\times N}$ is the evolution matrix, $\mathbf{B} \in \mathbb{R}^{N\times 1}$
and 
$\mathbf{C} \in \mathbb{R}^{1\times N} $ are the projection parameters.

\begin{align}
h^{'}{t} &= \mathbf{A}h(t) + \mathbf{B}x(t),\\
y(t) &= \mathbf{C}h(t)
\end{align}

\begin{figure}
\centering
\includegraphics[width=\linewidth]{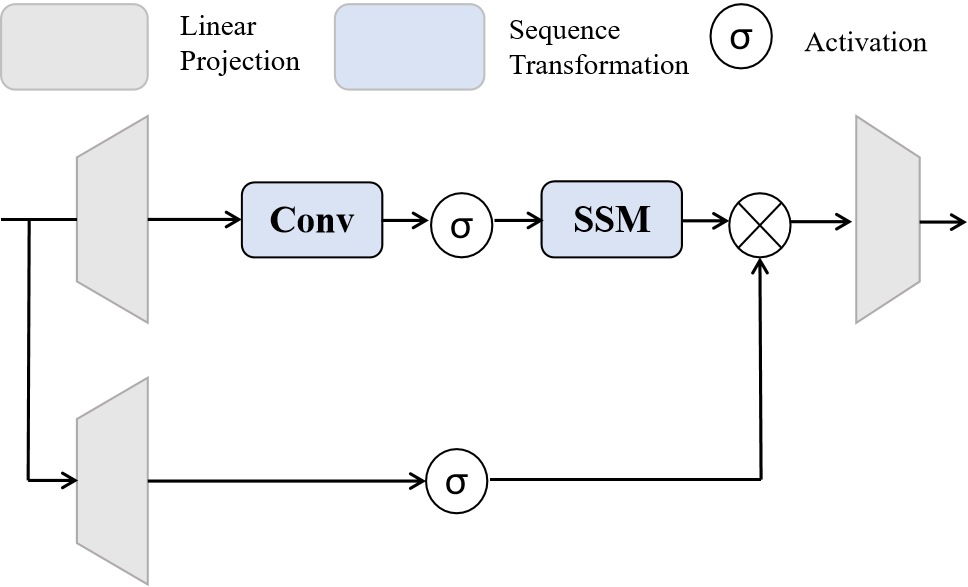}   
    \caption{\textbf{Mamba blocks for 1D sequence.}  }
\label{fig:mamba}
\end{figure}

The S4 and Mamba are the discrete versions of the continuous system, which include a timescale parameter $\Delta$ to
transform the continuous parameters ${\mathbf{A}},{\mathbf{B}}$ to their discrete counterparts
$\overline{\mathbf{A}}, \overline{\mathbf{B}}$.
The transformation formally uses a zero-order hold (ZOH) method, 
after the discretization of $\overline{\mathbf{A}}, \overline{\mathbf{B}}$, we get the discretized version as follows:
\begin{align}
\overline{\mathbf{A}} & = \exp (\boldsymbol{\Delta} \mathbf{A}), \\
\overline{\mathbf{B}} & = (\boldsymbol{\Delta} \mathbf{A})^{-1}(\exp (\boldsymbol{\Delta} \mathbf{A})-\mathbf{I}) \cdot \boldsymbol{\Delta} \mathbf{B} \\
h_{t} & = \overline{\mathbf{A}} h_{t-1}+\overline{\mathbf{B}} x_{t}, \\
y_{t} & = \mathbf{C} h_{t} .
\end{align}
And finally, the models output through a global convolution.
Mamba designs a Selective Scan Mechanism (S6) as its core SSM operator as shown in Fig~\ref{fig:mamba}.


\begin{table*}[h]
\caption{\textbf{Ablation for other C-Mamba designs.} \textbf{B} denotes bi-directional Mamba, \textbf{T} denotes temporal embedding, \textbf{L} denotes low resolution branch, \textbf{H} denotes high resolution branch.
}

\centering
\resizebox{0.8\linewidth}{!}{ 

    \begin{tabular}{cccc|cccc|cccc} 
    \hline
    \multicolumn{4}{c|}{\textbf{Dataset}} 
    &\multicolumn{4}{c|}{\textbf{DRAMA}} & \multicolumn{4}{c}{\textbf{BDD-X}}  \\
    \hline
    \multirow{2}{*}{B} &\multirow{2}{*}{T} &\multirow{2}{*}{L} &\multirow{2}{*}{H} 
    & \multicolumn{2}{c}{Caption} & \multicolumn{2}{c|}{Detection} & \multicolumn{2}{c}{Description} &  \multicolumn{2}{c}{Justification}           \\
    & & & &   \textbf{B4} $\uparrow$ &  \textbf{C} $\uparrow$ & \textbf{mIoU}$\uparrow$  & \textbf{Acc}$\uparrow$ &\textbf{B4} $\uparrow$ &  \textbf{C} $\uparrow$ &\textbf{B4} $\uparrow$  & \textbf{C} $\uparrow$ \\
    \hline

    \XSolidBrush
    & \XSolidBrush
    & \XSolidBrush
    & \XSolidBrush
    & 0.513 & 3.002  & 0.640 & 0.708 & 0.244 & 1.349 & 0.072 & 0.709  \\
    
      \XSolidBrush
    & \Checkmark
    & \Checkmark
    & \Checkmark
    & 0.531 & 3.135 & 0.665 & 0.752 & 0.292 & 1.961 & 0.087 & 0.936  \\

     \Checkmark
    & \XSolidBrush
    & \Checkmark
    & \Checkmark
    & 0.534 & 3.148 & 0.664 & 0.746 & 0.291 & 1.963 & 0.086 & 0.930   \\

      \Checkmark
    & \Checkmark
    & \XSolidBrush
    & \Checkmark
    & 0.536  & 3.227 & 0.669  & 0.756 & 0.298  & 1.991 & 0.088 & 0.949\\
    
      \Checkmark
    & \Checkmark
    & \Checkmark
    & \XSolidBrush
    & 0.530  & 3.160 &  0.664  & 0.745 & 0.290  & 1.973 & 0.084  & 0.919 \\

      \Checkmark
    & \Checkmark
    & \Checkmark
    & \Checkmark
    & 0.534  & 3.176 &  0.667  & 0.754 & 0.301 & 1.996 & 0.088  & 0.944 \\
     \hline  
    \end{tabular}
    }

\label{tab:design}      
\end{table*}

\section{Additional Ablations}
After the choice of Mamba-based modules, we then evaluate some designs in our hierarchical C-Mamba structure, including the leverage of bidirectional Mamba(Bi-Mamba), temporal embedding, as well as high and low temporal resolution designs.
As shown in Table \ref{tab:design}, 
(1)we first evaluate the influence of Bi-Mamba, which enhances the capacity for spatially-aware processing.
The results demonstrate a comprehensive performance decline when using only unidirectional Mamba.
(2)Then we explore the effect of temporal embedding, which is added to the Joint ST Mamba for better distinction of frames.
The model with PE has an improvement of mIoU for detection, and for caption on average.
(3)Finally, for our high and low temporal resolution designs, we ablate with only high or low branch to show the difference.
The table shows that high temporal resolution branch has slight performance advantage than the low branch on the whole,
while for some tasks, 
the combination of them yield better performance, 
this proves that different temporal resolution branches are complementary to some extent under specific circumstances, and a proper combination could result in more robust results.

Then we consider removing any one component and only use the combination of two Mamba modules in Table \ref{tab:two}.
The results indicate that the usage of two modules show some advantages in some specific metrics, but still they may not achieve the best performance in other metrics.
and with all of them, we get a unified framework for various tasks to get the optimal results.

For the comparison with CNN-style models for video processing, we choose the most representative model I3D\cite{i3d} R50 for instance. In order to use 3d kernel, we reshape the visual features to shape of $H \times W \times T$ and then process the video with I3D. It has about 28M parameters, comparable with Mamba, but the inference flops is nearly 200 GFLOPs, which is much larger than that of Mamba. In general, Mamba modules have lower training and inference FLOPs than 3D Spatio-Temporal CNN-style. And the performance of I3D on Drama is 0.520, 3.142 for B4, C and 66.3\% for mIoU, it is also inferior to our method.

\begin{table}[t]
\caption{\textbf{Ablation for only using two of the modules in the C-Mamba.}
}
\centering
\resizebox{\linewidth}{!}{ 
    \begin{tabular}{l|cccc} 
    \hline
    \multirow{2}{*}{\textbf{Module}}  & \multicolumn{2}{c}{Caption} &\multicolumn{2}{c}{Detection}        \\
     & \textbf{B4} $\uparrow$  & \textbf{C} $\uparrow$ & \textbf{mIoU}$\uparrow$ & \textbf{Acc}$\uparrow$    \\
    \hline

    T+JST Mamba & 0.535  & 3.206 & 66.6 & 75.0 
    \\
    T+DST Mamba & 0.533  & 3.196 & 66.7 & 75.2 
    \\
    DST+JST Mamba & 0.536  & 3.227 & 66.4 & 74.9
    \\
   
    \hline  
    \end{tabular}
    }
\label{tab:two}      
\end{table}

\begin{figure}
\centering
\includegraphics[width=\linewidth]{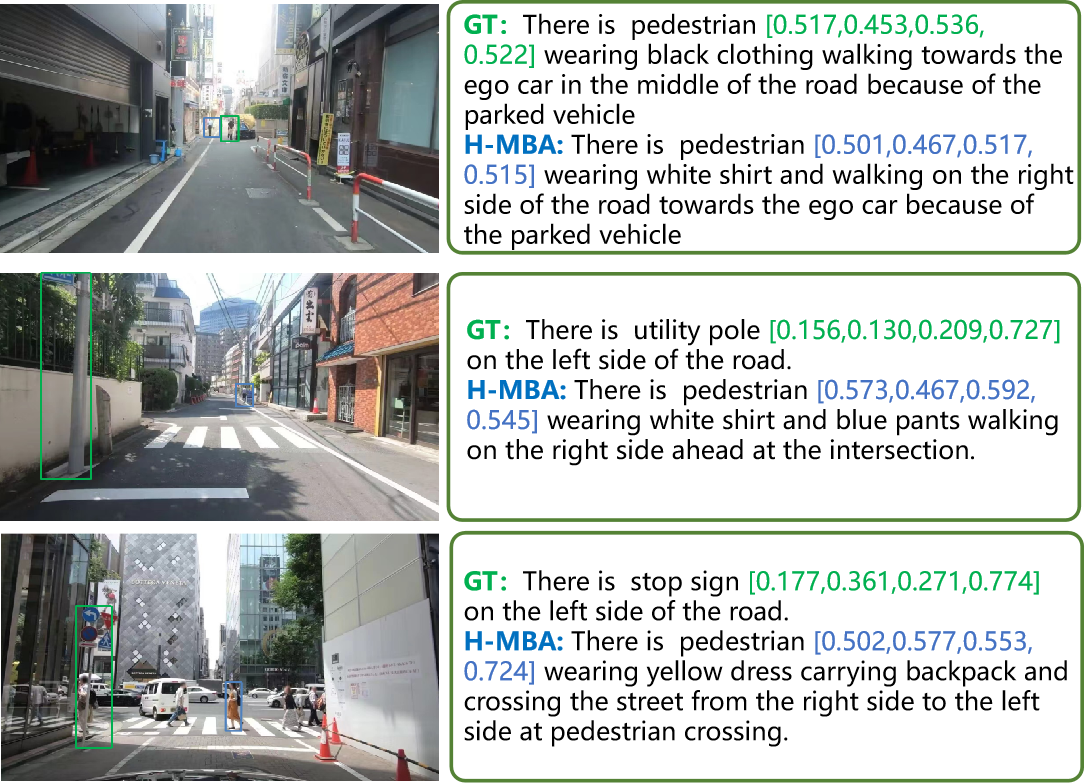}   
    \caption{\textbf{Failure examples of the model.}  }
\label{fig:fail}
\end{figure}

\section{Additional Visualization}
In this section, 
we provide some cases where our model produces results different from the ground truth answers.
As shown in the Fig \ref{fig:fail}, when faced with multiple potential risk objects, H-MBA may focus on other object different from the GT.
For example, the answer puts more attention on the pedestrian and it's somewhat reasonable, though the GT is the stop sign.
It remands us of two things, one is that we should enable the model to output coordinates for multiple risk objects, providing greater flexibility.
Another is that the model should logically assess the risk ratings of multiple objects according to traffic rules, enabling better handling of situations with multiple objects and complicated scenes.

\begin{figure*}
\centering
\includegraphics[width=0.8\linewidth]{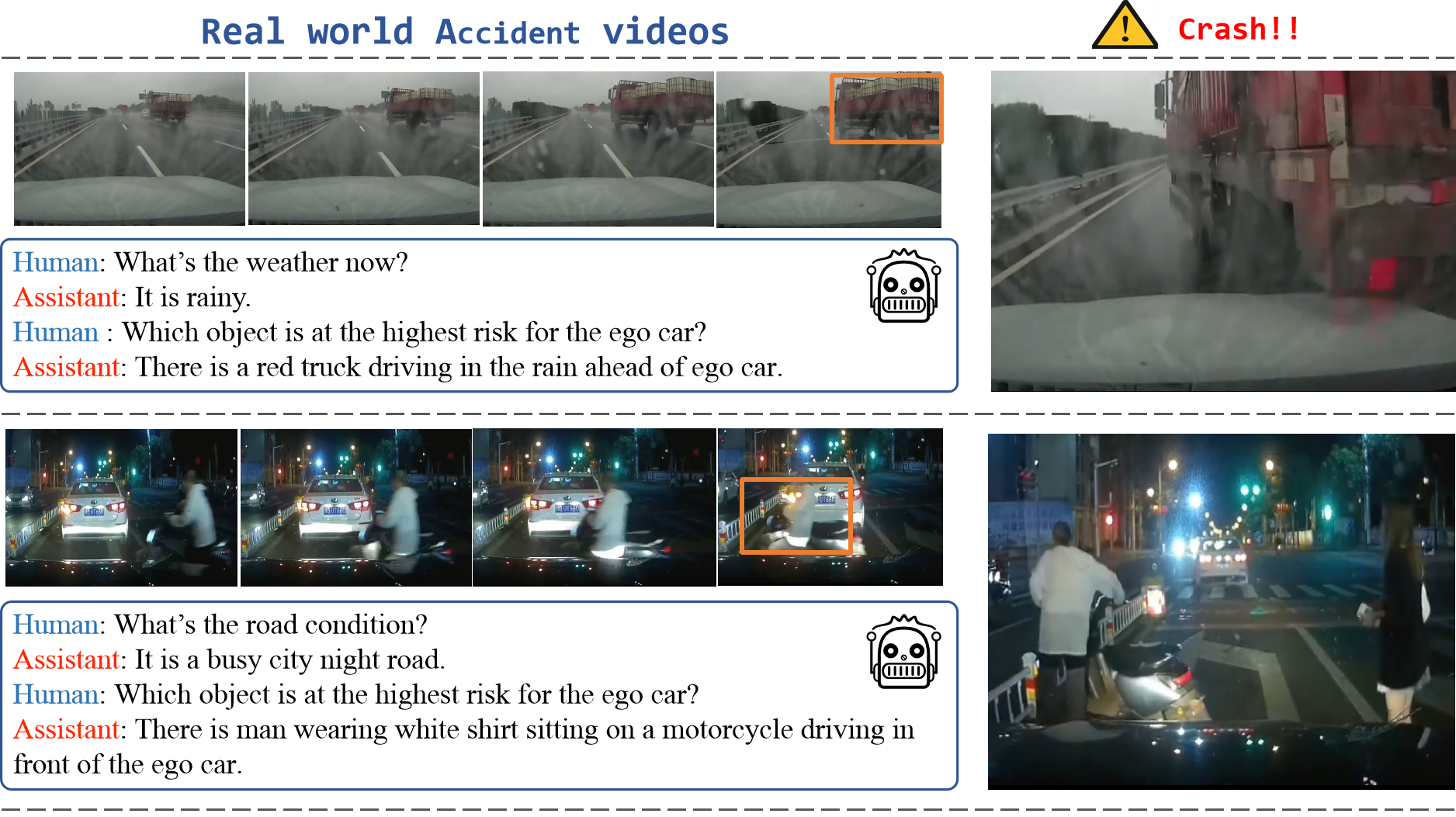}   
    \caption{\textbf{Visualization example of real world accident videos.} Our model can accurately identify the environmental conditions such as rainy weather and nighttime and provide precise risk warnings with bounding boxes, showing great flexibility and generalization ability. }
\label{fig:real}
\end{figure*}

Furthermore, 
we also collected some real traffic accident videos from the Internet to validate whether our model can generalize to real-world scenarios and correctly identify impending risks in Fig \ref{fig:real}. 
These videos are all captured by dashcams, which have significant differences from the training dataset.
We select the video clip several seconds before the accident as the input, our model can accurately identify the environmental conditions such as rainy weather and nighttime and provide precise risk warnings with bounding boxes.
It demonstrates excellent generalization and the potential for practical application of risk warning of our model to enhance the safety of autonomous driving.

\section{Limitations}
As a large-scale pre-trained model, the model's output results are greatly influenced by the training data. However, we are limited to conducting experiments only on existing publicly available datasets, which imposes constraints on verifying other aspects of autonomous driving tasks.
For example, there may be over one risk object to be notices on the road, and Shikra is free to output multi objects with bounding boxes, but in DRAMA only one is required.
Further more, the current model gives relatively intuitive judgments based on the video context, without consideration of traffic rules.
To fully leverage the reasoning capabilities of large language models, the chain of thought process is required to enhance its logical clarity, that's also our future research direction.


\bibliography{aaai25.bib}